%% file: main.tex
\documentclass[11pt]{article}
\PassOptionsToPackage{bookmarks=false}{hyperref}
\ifdefined\NAACLCameraReady
  \usepackage[final]{acl}
\else
  \usepackage[preprint]{acl}
\fi
\usepackage{fontspec}
\usepackage{amsmath,booktabs,tabularx,array,multirow,graphicx,xurl}
\usepackage{float}
\usepackage{stfloats} 
\usepackage{needspace}
\usepackage{tikz,pgfplots}
\usetikzlibrary{arrows.meta,positioning,calc,shapes.multipart}
\pgfplotsset{compat=1.18}
\usepackage{listings}
\definecolor{ink}{HTML}{253D53}
\definecolor{teal}{HTML}{187F88}
\definecolor{ochre}{HTML}{BE8234}
\definecolor{bluewash}{HTML}{EEF3F8}
\definecolor{tealwash}{HTML}{EDF6F4}
\newcolumntype{Y}{>{\raggedright\arraybackslash}X}
\renewcommand{\abstractname}{Abstract}

\setcitestyle{authoryear,round,semicolon}
\newlength{\papertablesep}
\hypersetup{pdftitle={Description-Guided Intermediate Representations: Synthesizing Schemas for Smart Home Automation},
  pdfauthor={Hyeongnam Jang}}
\title{Description-Guided Intermediate Representations:\\Synthesizing Schemas for Smart Home Automation}

\author{Hyeongnam Jang \\
  Samsung Electronics \\
  \texttt{hypeboy.jang@samsung.com}}

\begin{document}
\maketitle

\begin{abstract}
SmartThings, Samsung Electronics' smart home platform with hundreds of millions of registered users, provides automations called routines that execute actions when specified conditions are met. We develop a library that converts natural-language requests into these routines. We automate its design with a high-performance LLM, iteratively refining it for a separate LLM that generates routines in the service. The platform's JSON requires complex nesting and explicit value types that LLMs often generate incorrectly. An intermediate representation (IR) can express behavior more simply. However, designing its schema from routine JSON can reproduce that complexity, leaving LLMs with complex structures to generate. We propose Description-Guided IR. We generate natural-language descriptions from requests and paired routine examples, preserving behavior while excluding the platform's JSON structure. Requests and descriptions guide automatic design of the IR's format and semantics. Conversion code transforms LLM-generated IR into final routines. RoutineGen Refiner implements this process, evaluating routines from the service LLM and incorporating failure analysis and developer review. The library is deployed in SmartThings Pet Care Assistant. With the same service LLM, mean offline accuracy was 98\% with Description-Guided IR, 95\% with IR designed from routine JSON, and 85\% with direct generation without IR.
\end{abstract}

\section{Introduction}
\label{sec:introduction}
SmartThings routines execute actions when specified conditions are met. Generating them from natural-language requests requires preserving user intent while meeting platform and device constraints. We use a high-performance LLM to automatically design generation libraries and refine them for a separate LLM that generates routines in the service (the \emph{service LLM}).

The platform's routine format is difficult for LLMs to generate directly because even simple behaviors require nested structures and explicit value types. An \emph{intermediate representation} (IR) lets the service LLM express the requested behavior in a simpler structured form, while a translator constructs the required routine JSON.

This separation depends on the IR \emph{schema}, which defines its structure and meaning. Designing it from routine JSON can reproduce its nesting and type wrappers, leaving the service LLM to generate similarly complex structures. \emph{Description-Guided IR} instead uses requests and natural-language descriptions of paired routines. The descriptions preserve behavior but omit JSON structure from schema-design inputs.

A library that works with the high-performance LLM may fail with the service LLM. \emph{RoutineGen Refiner} builds a library containing an IR schema, generation instructions and examples, and a translator, then evaluates routines generated by the service LLM. In an automated loop, the high-performance LLM analyzes failures and revises the instructions, examples, or translator. Development also exposed erroneous expected outputs and conflicting interpretations of device specifications and service requirements. These issues require more than library edits: in an outer loop, developers review AI proposals, correct data or shared rules, and guide the next design and refinement cycle.

We contribute automated IR design from behavior descriptions and a development process that uses service-model evaluation and developer judgment to refine libraries, data, and routine requirements. The resulting library is deployed in SmartThings Pet Care Assistant. Offline comparisons examine representation choices and the role of service-model outputs in refinement.

\begin{figure*}[t]
\centering
\input{figures/semantic_ir_payload_example}
\caption{Illustrative comparison for a no-motion routine. The routine payload and Baseline IR nest actions within the daily execution limit; Description-Guided IR separates the limit from the action sequence. Colors mark corresponding semantics; ellipses omit metadata and device details.}
\label{fig:semantic-ir-example}
\end{figure*}
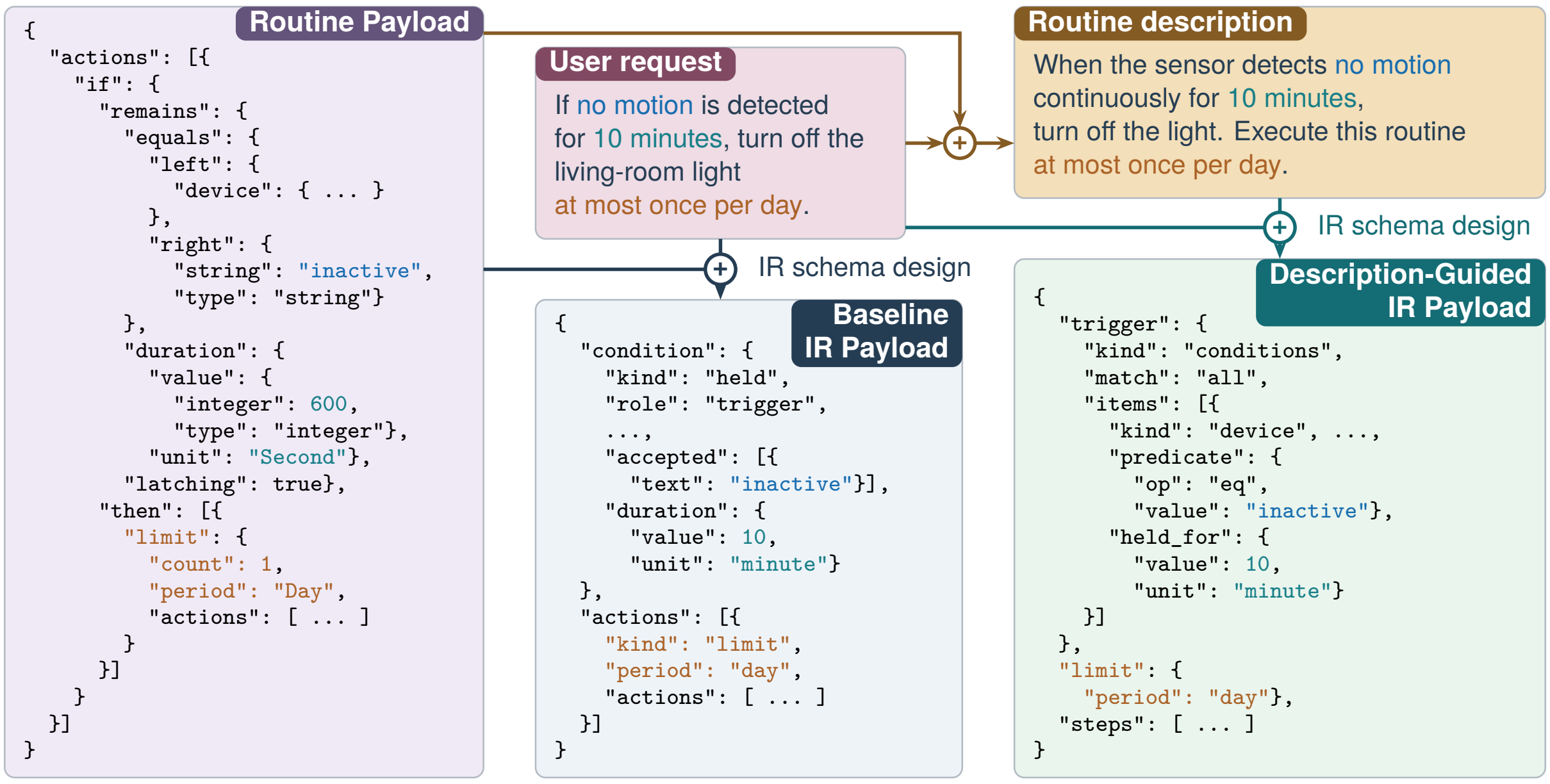

\section{Related Work}
\label{sec:related}
\noindent\textbf{Smart Home Automation.} Sasha \citep{king2024sasha} builds plans from user goals, while \mbox{RuleBot++} \citep{gallo2024rulebot} creates automations through conversation. SimuHome \citep{seo2026simuhome} evaluates agents under changing time and environments. We focus on developing reusable generation libraries for SmartThings, including the representation and translation needed to construct routines.

\noindent\textbf{IR Design.} IR-based generation separates structured output from final-format compilation \citep{pan2025instance}, while Doc2Spec \citep{xia2026doc2spec} induces specification grammars from natural-language rules. We study schema design from behavior descriptions derived from request--routine pairs, excluding the platform's JSON structure from the design inputs.

\noindent\textbf{Refinement and Developer Review.} DSPy \citep{khattab2024dspy} and AFlow \citep{zhang2025aflow} optimize model pipelines and workflows using evaluation feedback. EvalGen \citep{shankar2024evalgen} aligns evaluation with human judgments; AI Chains \citep{wu2022aichains} exposes intermediate outputs for inspection and editing. We connect automated library refinement with developer decisions about data and routine requirements, addressing failures that library edits alone cannot resolve.

\section{Description-Guided IR}
\label{sec:semantic-projection}
\begin{figure*}[t]
\centering
\scalebox{0.88}{\input{figures/refiner_two_loops}}
\caption{The refiner (left) uses the analysis LLM for bootstrapping and refines the library. Service execution (right) shows the main routine-generation path in Pet Care Assistant on SmartThings.}
\label{fig:refiner-two-loops}
\end{figure*}
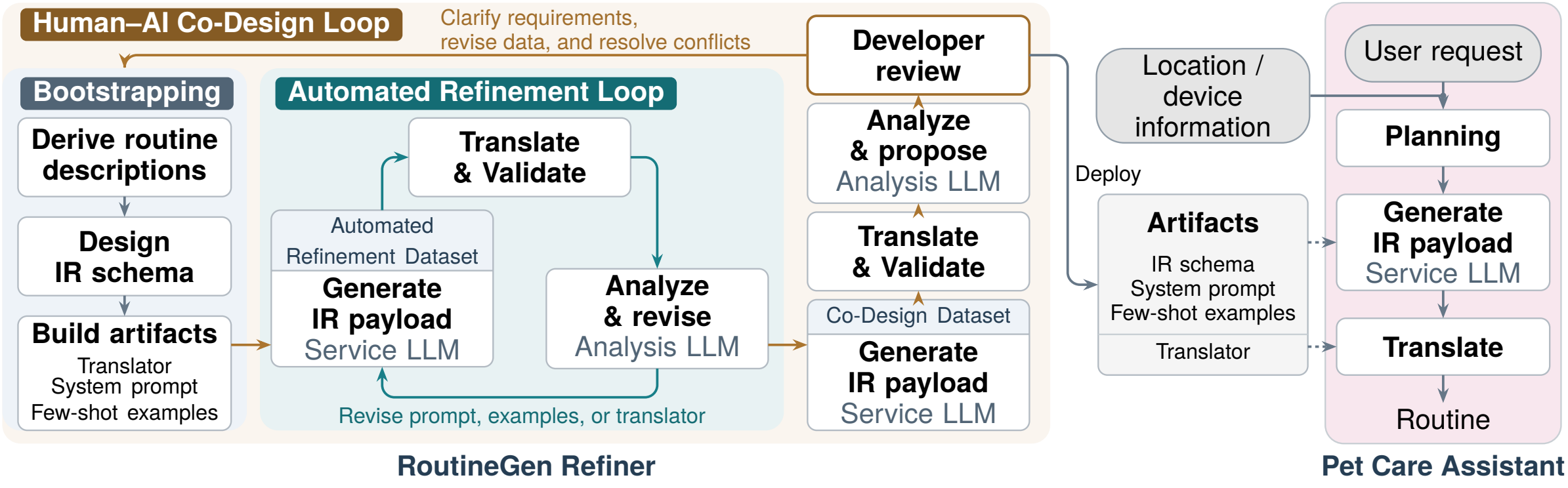

Description-Guided IR aims to express routine behavior without reproducing the platform's routine format.

\subsection{Routine Schema}
\label{sec:routine-schema}
A schema defines a data structure and the types and constraints of its fields. A payload contains concrete values organized according to that schema.

The SmartThings routine schema represents conditions, schedules, and actions in JSON \citep{smartthings-rules}. Device commands identify targets, capabilities, commands, and arguments; control structures specify conditional execution and action order, with explicit value types and units.

The routine payload in Figure~\ref{fig:semantic-ir-example} uses explicitly typed values and nested structures for a state-duration check and an execution limit.

\subsection{Routine Descriptions}
\label{sec:routine-descriptions}
A user request alone may omit details that the routine must represent, while designing an IR schema directly from a routine payload may reproduce that payload's structure. We therefore generate a routine description from each user request and its paired routine payload. We use these requests and descriptions together to design the IR schema. During description generation, we instruct the model to preserve condition roles, logical scope, boundary values, targets, units, ordering, and delays while omitting the routine format's field names, nesting, and type annotations (Appendix~\ref{sec:description-details}).

The illustrative description in Figure~\ref{fig:semantic-ir-example} preserves the continuous no-motion condition, light-off action, and daily limit without the payload's wrappers.

\subsection{IR Schema}
\label{sec:ir-schema}
The IR schema defines the structure of the IR payloads generated by the service LLM and the meaning of each field. The LLM can focus on expressing routine semantics, such as conditions, schedules, and action order, while the translator constructs the final routine payload according to the routine specification.

Like \citet{pan2025instance}, we separate IR payload generation from construction of the final format. Their approach supplies a predefined JSON template, whereas our IR schema is synthesized from routine descriptions.

We reuse capability, command, and argument information from existing device specifications, avoiding IR schemas for numerous device types and extensions for new types or capabilities.

For comparison, \emph{Baseline IR} uses a schema designed from user requests, paired routine payloads, and the routine specification. Both IR methods automatically design a schema in each independent library-generation run, so field names and nesting can vary across libraries. Figure~\ref{fig:semantic-ir-example} shows example payloads from one library of each method.

In this example, Baseline IR nests device actions within the execution limit. Description-Guided IR places \texttt{limit} and \texttt{steps} at the same level. The service LLM expresses the limit and actions, while the translator constructs the final routine's nesting.

\section{RoutineGen Refiner}
\label{sec:framework}
Implemented with LangGraph \citep{langgraph}, the refiner coordinates bootstrapping and two nested loops (Figure~\ref{fig:refiner-two-loops}). The Automated Refinement Loop revises the library using validation results. The Human--AI Co-Design Loop uses developer review to revise data and requirements before the next design and refinement cycle.

The high-performance LLM (labeled ``Analysis LLM'' in Figure~\ref{fig:refiner-two-loops}) performs bootstrapping, failure analysis, and revision. Validation uses routines generated by the service LLM to expose failures that the high-performance model might miss. Neither loop updates model weights.

\subsection{Datasets}
\label{sec:data-construction}
\label{sec:evaluation-datasets}

\begin{table}[!t]
\centering\small
\captionsetup{position=bottom,skip=\papertablesep}

\setlength{\tabcolsep}{6pt}
\renewcommand{\arraystretch}{1.12}
\begin{tabularx}{\columnwidth}{@{}Xrrr@{}}
\toprule
Dataset & Generation & Refusal & Total \\
\midrule
Automated Refinement & 45 & 25 & 70 \\
Co-Design & 49 & 22 & 71 \\
Final Evaluation & 78 & 22 & 100 \\
\bottomrule
\end{tabularx}
\caption{Dataset composition by expected outcome. Counts denote requests.}
\label{tab:datasets}
\end{table}

\paragraph{Construction}
Privacy and other constraints precluded production user data. We used an LLM to create cases from the routine specification and platform/device constraints. The two development datasets were refined alongside the library. We reviewed the expected outputs for correctness. Each case includes a request, an expected routine or refusal, and location-specific device specifications. \emph{Generation} cases require supported behavior using available devices; \emph{refusal} cases involve unavailable devices or capabilities, unsupported behavior, or non-routine requests. Appendix~\ref{sec:dataset-details} describes environments and examples.

\paragraph{Dataset Roles}
The Automated Refinement Dataset supports bootstrapping, refinement, and candidate selection; the Co-Design Dataset informs developer review after the automated loop. All reported accuracies use the separate Final Evaluation Dataset, reserved for offline evaluation (Section~\ref{sec:evaluation}). We verified that no identical cases occur in both the development datasets and the Final Evaluation Dataset. Semantically similar requests may occur across these datasets; the split does not require disjoint routine behaviors.

\paragraph{Coverage and Size}
These datasets support library refinement rather than model-weight training. The 70-case Automated Refinement and 71-case Co-Design Datasets are each designed to cover all 10 action types and eight condition operators supported by the service (Appendix~\ref{sec:development-coverage}).

\subsection{Validation Criteria}
\label{sec:validation-criteria}
Generation cases require an exact match of the JSON structure and values between the final routine payload and expected output, with predefined exceptions detailed in Appendix~\ref{sec:validation-details}. Refusal cases require a refusal and a nonempty reason.

\subsection{Bootstrapping}
\label{sec:bootstrapping}
Doc2Spec \citep{xia2026doc2spec} induces a specification grammar from existing natural-language rules. In contrast, the refiner generates descriptions from user requests and paired routine payloads, omitting the payloads' original structure. For Description-Guided IR, the high-performance LLM uses these requests, descriptions, and general semantic requirements to design the IR schema. It does not receive routine payloads or routine specification documents at this stage. Both IR methods receive the same predefined requirements for condition roles, units, ordering, and refusals, including behaviors rare or absent in the development cases. We check whether the schema can express the required semantics and distinguish different behaviors. We then provide routine payloads, the routine specification, and device specifications as additional inputs to generate the translator. We reverse-convert routine payloads into IR payloads and translate them back into routines to check for information loss. Semantic review may lead to revisions of field meanings or the schema structure before the system prompt and few-shot examples are prepared. Appendix~\ref{sec:method-details} details these checks and revisions.

\subsection{Automated Refinement Loop}
\label{sec:inner-refinement}
Like DSPy \citep{khattab2024dspy} and AFlow \citep{zhang2025aflow}, the Automated Refinement Loop uses evaluation feedback to revise the system prompt, few-shot examples, or translator while keeping the bootstrapped IR schema fixed.

The service LLM generates IR payloads for the Automated Refinement Dataset using the current library and device information. Translated routines are compared with expected outputs.

The high-performance LLM analyzes failures using user requests, expected outputs, generated IR payloads, and translation results, then selects one artifact to revise. After revision, the refiner revalidates the entire Automated Refinement Dataset with the same service LLM, including previously passing cases. At termination, the candidate with the most passing cases is selected.

\subsection{Human--AI Co-Design Loop}
\label{sec:codesign}
After automated refinement, the refiner validates the selected library on the Co-Design Dataset using the service LLM. The developer reviews AI analyses and proposals from both stages against user requests, expected and generated routines, device and routine specifications, and service requirements covering request interpretation, supported behavior, and expected outputs.

Conflicts among these sources caused repeated revisions. The developer may adopt, adapt, or replace proposals after considering service inputs and effects on other cases. Decisions update data, shared rules, or implementation and guide the next bootstrapping and refinement cycle.

\paragraph{Which temperature unit should apply?}
\label{sec:temperature-case}
For a request with a threshold of ``25 degrees,'' the AI proposed converting 25 to 77 because the location's display setting was Fahrenheit. We distinguished this setting from the unit defined by the device specification. We established a shared rule: values without an explicit unit follow the device specification, and conversion applies only when the user explicitly specifies a different unit.

\paragraph{Which device does the request reference?}
\label{sec:device-reference-case}
A case asked to turn off a light after ten minutes without motion, at most once a day. The request and expected routine mistakenly referenced a dehumidifier as the motion sensor. The AI proposed using the living-room motion sensor's identifier instead. We adopted this proposal and corrected both the request and expected routine.

Developer review thus corrected expected outputs, which the automated loop cannot edit, and replaced a proposal conflicting with device specifications. Like EvalGen \citep{shankar2024evalgen}, we incorporate developer judgment into evaluation.

\section{Integration into SmartThings Pet Care Assistant}
\label{sec:service-integration}
Our library is deployed in SmartThings Pet Care Assistant, launched in September 2026. The assistant plans pet care using the devices available at each location. To avoid the cost and resource demands of an always-running agent monitoring and controlling devices, we use the existing routine system for execution. The assistant receives device context during planning to constrain plans to available devices. The service LLM uses the deployed schema, prompt, and examples to generate IR payloads from the plan and summarized context; the translator checks detailed device specifications and constructs routines (Appendix~\ref{sec:translator-construction}).

If routine generation refuses a plan, the assistant replans once with the refusal reason. If problems persist, only the affected routine is omitted; the remaining routines are generated. Users register the routines by selecting ``Start care,'' after which the existing routine system handles execution. This separates AI planning from persistent execution, bounds replanning, and prevents an individual routine's failure from blocking the entire plan. Figure~\ref{fig:refiner-two-loops} shows the main generation path.

\section{Offline Evaluation}
\label{sec:evaluation}
\label{sec:eval-conditions}
This section evaluates the routine-generation performance of libraries developed with the refiner, using the Final Evaluation Dataset described in Section~\ref{sec:data-construction}. The effects of the Human--AI Co-Design Loop and performance in service operation are outside the scope of this offline evaluation.

\subsection{Models}
Bootstrapping, analysis, and refinement used Claude Opus 5.5 (hereafter Opus) \citep{anthropic-opus55}; detailed task settings are listed in Appendix~\ref{sec:design-model-settings}. The reference service model was Gemini 3.5 Flash-Lite (hereafter Flash-Lite) with medium reasoning \citep{google-gemini35-flashlite}. The alternatives were Gemini 3.6 Flash (hereafter Flash) with low and minimal reasoning \citep{google-gemini36-flash}. We write each combination as model/level, e.g., Flash-Lite/medium, using the provider's reasoning levels \citep{google-thinking}.

\subsection{Method Comparison}
\paragraph{Direct Generation}
The service LLM generates routine payloads without an IR schema or translator. Its system prompt and few-shot examples are refined on the Automated Refinement Dataset.

\paragraph{Baseline IR}
The IR schema is designed from requests, paired routine payloads, and the routine specification, exposing payload field names, nesting, and typed values to the design LLM.

\paragraph{Description-Guided IR}
The IR schema is designed using requests and paired natural-language routine descriptions, preserving behavior while omitting the original payload structure (Section~\ref{sec:routine-descriptions}). The schema-design LLM does not receive routine payloads.

Both IR methods automatically design schemas and use the service LLM to generate IR payloads for translation into final routines. They differ in the inputs supplied during schema design. Figure~\ref{fig:semantic-ir-example} compares their IR payloads for the same request.

\subsection{Experimental Settings}
\label{sec:experimental-settings}
We evaluated all three methods with Flash-Lite/medium, Flash/low, and Flash/minimal. Each combination had three independent design and refinement runs (27 total). Accuracy is the fraction of requests that pass validation. Latency measures per-request routine generation, including model-call retries, translation, and postprocessing, but excluding validation against expected outputs and library development. It excludes assistant planning and routine registration. Retries apply only to model-call failures, not incorrect generated outputs. Generation settings, few-shot example counts, and stopping criteria appear in Appendix~\ref{sec:eval-details}.

With a common cap of 60 revised-candidate evaluations, actual totals ranged from 4 to 29 including the initial evaluation; none exhausted the cap. Runs ended when all development cases passed, progress stagnated, or the analysis model explicitly decided to stop. Appendix~\ref{sec:recorded-budgets} reports realized budgets and termination reasons.

\subsection{Evaluation Results}
Description-Guided IR exceeded Direct Generation in mean accuracy by 4.67--13.33 percentage points (Table~\ref{tab:mainresults}). Its largest gain over Baseline IR was with Flash-Lite/medium: 98.00\% versus 95.00\%. With Flash/low and Flash/minimal, gains were under one point and run-level accuracies overlapped (Table~\ref{tab:accuracy-runs}). Description-Guided IR had higher latency than Baseline IR in all three configurations. Each run develops prompts, examples, and, for IR methods, a translator. These results compare complete libraries rather than isolating the effect of routine descriptions.

Service libraries need both high accuracy and consistent quality across repeated development runs. Large performance differences between runs make deployment quality harder to predict and manage. Across all three tested model settings, Description-Guided IR achieved higher mean accuracy and narrower observed accuracy ranges than Direct Generation, showing more consistent final performance across the three independent development runs per setting.

For a request to set an air conditioner to dry mode daily at 17:00, one Baseline IR library required \texttt{\{"text":"dry"\}}, but the service LLM generated \texttt{"dry"}, causing translation to fail. A Description-Guided IR library allowed this scalar value, translated it into the routine, and passed evaluation for the same request. Appendix~\ref{sec:recorded-value-case} provides excerpts of the recorded schemas, IR payloads, and translation outcomes.

\begin{table}[!t]
\centering\fontsize{8.5}{10}\selectfont
\captionsetup{position=bottom,skip=\papertablesep}

\setlength{\tabcolsep}{1.5pt}
\renewcommand{\arraystretch}{1.12}
\begin{tabularx}{\columnwidth}{@{}l>{\hspace{0.25em}}Yrr@{}}
\toprule
Model & Method & \shortstack{Accuracy (\%)\\Mean [runs 1--3]} & \shortstack{Latency\\(s)} \\
\midrule
\multirow{3}{*}{\shortstack[l]{Flash-Lite\\/medium}} & Direct Generation & 84.67 [82, 86, 86] & 6.13 \\
& Baseline IR & 95.00 [92, 94, 99] & 4.59 \\
& Description-Guided IR & \textbf{98.00} [97, 98, 99] & 4.68 \\
\midrule
\multirow{3}{*}{\shortstack[l]{Flash\\/low}} & Direct Generation & 86.33 [83, 85, 91] & 5.12 \\
& Baseline IR & 95.67 [95, 96, 96] & 3.67 \\
& Description-Guided IR & 96.00 [95, 95, 98] & 4.23 \\
\midrule
\multirow{3}{*}{\shortstack[l]{Flash\\/minimal}} & Direct Generation & 87.33 [84, 87, 91] & 1.79 \\
& Baseline IR & 91.33 [91, 91, 92] & 1.72 \\
& Description-Guided IR & 92.00 [91, 92, 93] & 1.78 \\
\bottomrule
\end{tabularx}
\caption{Mean accuracy and mean latency across independently developed libraries per configuration.}
\label{tab:mainresults}
\end{table}

Among the tested Description-Guided IR settings, Flash-Lite/medium achieved the highest accuracy, with latency comparable to Flash/low. Its estimated regular-price request costs were 66--69\% below the two Flash configurations (Appendix~\ref{sec:cost-details}). Per-run results and token usage appear in Appendix~\ref{sec:details}.

For Description-Guided IR with Flash-Lite/medium, accuracy ranged from 44\% to 95\% after bootstrapping and rose to 97--99\% after automated refinement (Appendix~\ref{sec:initial-improvement-details}). All three runs reached high final accuracy despite differing initial quality. In these runs, refinement helped compensate for variation in the initial libraries.

\subsection{Role of the Service LLM in Refinement}
\label{sec:cross-model}
\begin{table}[!t]
\centering\small
\captionsetup{position=bottom,skip=\papertablesep}
\setlength{\tabcolsep}{4pt}
\begin{tabularx}{\columnwidth}{@{}lYrr@{}}
\toprule
Development & \shortstack[l]{Generation\\LLM} & Acc.\ (\%) & Latency (s) \\
\midrule
Opus only & Opus & 97.33 & 3.83 \\
Opus only & Flash-Lite & 93.33 & 4.70 \\
Opus + Flash-Lite & Flash-Lite & 98.00 & 4.68 \\
\bottomrule
\end{tabularx}
\caption{Mean performance by service LLM use during refinement. Generation LLM identifies the model used to generate IR payloads during final evaluation.}
\label{tab:cross-model}
\end{table}

With three Opus-only libraries held fixed, switching IR payload generation from Opus to Flash-Lite reduced accuracy by 4.00 percentage points. Independently developed libraries refined with Flash-Lite outputs scored 4.67 points higher with Flash-Lite (Table~\ref{tab:cross-model}). Like AFlow's evaluation with the execution model \citep{zhang2025aflow}, this supports validation with the intended service LLM, but introduces model dependence and maintenance when models change \citep{sculley2015technicaldebt}. Opus performed design, analysis, and revision in all conditions; Appendix~\ref{sec:cross-model-details} provides detailed results.

\section{Conclusion}
We proposed Description-Guided IR, designing schemas from natural-language descriptions that omit the platform's routine JSON structure and separating semantic expression from construction of the execution format. Libraries using this IR achieved higher offline accuracy than Direct Generation.

RoutineGen Refiner builds libraries through automated refinement and developer review. For Description-Guided IR with Flash-Lite/medium, refinement achieved consistently high accuracy despite variation in initial quality. Libraries developed only with a high-performance LLM were 4.67 percentage points less accurate with the service LLM than those refined using its outputs, supporting refinement with the intended model. Developer review resolved conflicts in requirements and data. The library is deployed in Pet Care Assistant on SmartThings. As the assistant expands to general routine generation, we plan to support modification of previously generated routines by defining input requirements and extending the datasets.

\label{page:main-end}

\section*{Limitations}
This paper presents our experience developing a routine generator deployed in Pet Care Assistant on SmartThings, supplemented with a limited set of additional experiments. Due to practical constraints, we did not evaluate generalization to other domains or unseen location environments. The Final Evaluation Dataset contains separate requests in specified environments, with semantically similar requests permitted across development and evaluation datasets.

Refusal evaluation checks whether a request is refused with a nonempty reason; it does not assess the correctness of that reason. Offline evaluation measures the routine-generation library, not the end-to-end assistant, including planning, refusal-guided replanning, routine registration, or effects on pet care.

We describe the Human--AI Co-Design Loop as development experience; its effectiveness is outside the scope of the quantitative evaluation. The reported scores do not establish the best performance each method could attain. Further human--AI co-design and adjustments for each service model may improve any of the methods, including Direct Generation and Baseline IR. We have not measured these additional gains or whether they would change the observed performance gaps.

\bibliography{references}
\clearpage
\appendix
\input{appendix}
\end{document}

%% file: figures/semantic_ir_payload_example.tex
\begingroup
\definecolor{motion}{HTML}{176DB0}
\definecolor{duration}{HTML}{187F88}
\definecolor{frequency}{HTML}{AB6227}
\definecolor{payloadbg}{HTML}{F0EAF7}
\definecolor{baselinebg}{HTML}{EEF3F8}
\definecolor{guidedbg}{HTML}{EAF5ED}
\definecolor{requestbg}{HTML}{EFDCE5}
\definecolor{descriptionbg}{HTML}{F3DFB9}
\definecolor{payloadtitle}{HTML}{64537A}
\definecolor{requesttitle}{HTML}{824963}
\lstset{basicstyle=\ttfamily\fontsize{8}{8}\selectfont,
  columns=fullflexible,keepspaces=true,breaklines=false,
  showstringspaces=false,frame=none,backgroundcolor={},
  xleftmargin=0pt,xrightmargin=0pt,aboveskip=0pt,belowskip=0pt,
  escapeinside={(*@}{@*)}}
\newsavebox{\routinePayloadCode}
\begin{lrbox}{\routinePayloadCode}\begin{minipage}{4.65cm}
\begin{lstlisting}
{
  "actions": [{
    "if": {
      "remains": {
        "equals": {
          "left": {
            "device": { ... }
          },
          "right": {
            "string": (*@\textcolor{motion}{"inactive"}@*),
            "type": "string"}
        },
        "duration": {
          "value": {
            "integer": (*@\textcolor{duration}{600}@*),
            "type": "integer"},
          "unit": (*@\textcolor{duration}{"Second"}@*)},
        "latching": true},
      "then": [{
        (*@\textcolor{frequency}{"limit"}@*): {
          (*@\textcolor{frequency}{"count": 1}@*),
          (*@\textcolor{frequency}{"period": "Day"}@*),
          "actions": [ ... ]
        }
      }]
    }
  }]
}
\end{lstlisting}
\end{minipage}\end{lrbox}
\newsavebox{\baselineIRCode}
\begin{lrbox}{\baselineIRCode}\begin{minipage}{4.10cm}
\begin{lstlisting}
{
  "condition": {
    "kind": "held",
    "role": "trigger",
    ...,
    "accepted": [{
      "text": (*@\textcolor{motion}{"inactive"}@*)}],
    "duration": {
      "value": (*@\textcolor{duration}{10}@*),
      "unit": (*@\textcolor{duration}{"minute"}@*)}
  },
  "actions": [{
    (*@\textcolor{frequency}{"kind": "limit"}@*),
    (*@\textcolor{frequency}{"period": "day"}@*),
    "actions": [ ... ]
  }]
}
\end{lstlisting}
\end{minipage}\end{lrbox}
\newsavebox{\guidedIRCode}
\begin{lrbox}{\guidedIRCode}\begin{minipage}{5.20cm}
\begin{lstlisting}
{
  "trigger": {
    "kind": "conditions",
    "match": "all",
    "items": [{
      "kind": "device", ...,
      "predicate": {
        "op": "eq",
        "value": (*@\textcolor{motion}{"inactive"}@*)},
      "held_for": {
        "value": (*@\textcolor{duration}{10}@*),
        "unit": (*@\textcolor{duration}{"minute"}@*)}
    }]
  },
  (*@\textcolor{frequency}{"limit"}@*): {
    (*@\textcolor{frequency}{"period": "day"}@*)},
  "steps": [ ... ]
}
\end{lstlisting}
\end{minipage}\end{lrbox}

\noindent\resizebox{\textwidth}{!}{%
\begin{tikzpicture}[x=1cm,y=1cm,
  card/.style={draw=ink!35,line width=.5pt,rounded corners=3pt,
    anchor=north west,minimum width=5.05cm,inner sep=0pt,outer sep=0pt},
  badge/.style={anchor=north west,font=\sffamily\bfseries\fontsize{8.5}{10}\selectfont,
    text=white,rounded corners=3pt,inner xsep=4pt,inner ysep=1pt,
    outer sep=0pt},
  prose/.style={anchor=north west,align=left,inner sep=0pt,outer sep=0pt,
    text=ink,font=\sffamily\fontsize{8}{10}\selectfont},
  flow/.style={draw=ink!70,line width=1.05pt},
  arrow/.style={flow,-{Stealth[length=1.9mm]}},
  flowlabel/.style={anchor=west,font=\sffamily\fontsize{8}{9.5}\selectfont,
    text=ink,fill=white,inner xsep=2pt,inner ysep=1pt}]

  \node[card,fill=payloadbg,
    minimum height=\dimexpr\ht\routinePayloadCode+\dp\routinePayloadCode+.26cm\relax]
    (payload) at (0,0) {};
  \node[badge,fill=payloadtitle,anchor=north east] at ($(payload.north east)+(0,0)$)
    {Routine Payload};
  \node[anchor=north west,inner sep=0pt,outer sep=0pt]
    at ($(payload.north west)+(.20,-.13)$) {\usebox{\routinePayloadCode}};

  \node[card,fill=requestbg,minimum width=3.90cm,minimum height=2.02cm] (request) at (5.60,-.43) {};
  \node[badge,fill=requesttitle] at ($(request.north west)+(0,0)$)
    {User request};
  \node[prose] at ($(request.north west)+(.20,-.50)$)
    {If \textcolor{motion}{no motion} is detected\\
     for \textcolor{duration}{10 minutes}, turn off the\\
     living-room light\\
     \textcolor{frequency}{at most once per day}.};

  \node[card,fill=descriptionbg,minimum width=5.60cm,minimum height=2.02cm] (description) at (10.65,0) {};
  \node[badge,fill=ochre!70!black] at ($(description.north west)+(0,0)$)
    {Routine description};
  \node[prose] at ($(description.north west)+(.20,-.50)$)
    {When the sensor detects \textcolor{motion}{no motion}\\
     continuously for \textcolor{duration}{10 minutes},\\
     turn off the light. Execute this routine\\
     \textcolor{frequency}{at most once per day}.};

  \node[card,fill=baselinebg,anchor=south west,minimum width=4.50cm,
    minimum height=\dimexpr\ht\baselineIRCode+\dp\baselineIRCode+.26cm\relax]
    (baseline) at (5.60,0 |- payload.south) {};
  \node[badge,fill=ink,anchor=north east,align=right]
    at ($(baseline.north east)+(0,0)$)
    {Baseline\\IR Payload};
  \node[anchor=north west,inner sep=0pt,outer sep=0pt]
    at ($(baseline.north west)+(.20,-.13)$) {\usebox{\baselineIRCode}};

  \path let \p1=($(request.south)-(baseline.north)$),
            \p2=($(description.south)-(payload.south)$) in
    node[card,fill=guidedbg,anchor=south west,minimum width=5.60cm,
      minimum height=\dimexpr\y2-\y1\relax]
      (guided) at (description.west |- payload.south) {};
  \node[badge,fill=teal!85!black,anchor=north east,align=right]
    at ($(guided.north east)+(0,0)$)
    {Description-Guided\\IR Payload};
  \node[anchor=south west,inner sep=0pt,outer sep=0pt]
    at ($(guided.south west)+(.20,.13)$) {\usebox{\guidedIRCode}};

  \tikzset{inputmerge/.style={circle,fill=white,line width=1.05pt,
    minimum size=.32cm,inner sep=0pt,outer sep=0pt,
    font=\sffamily\bfseries\fontsize{8.5}{8.5}\selectfont}}
  \node[inputmerge,draw=ochre!70!black,text=ochre!70!black]
    (descriptionjoin) at (10.075,0 |- request.east) {+};
  \draw[arrow,draw=ochre!70!black] (request.east) -- (descriptionjoin.west);
  \coordinate (descriptionrail) at ($(request.north)+(0,.15)$);
  \draw[arrow,draw=ochre!70!black] (payload.east |- descriptionrail)
    -- (descriptionjoin.north |- descriptionrail) -- (descriptionjoin.north);
  \draw[arrow,draw=ochre!70!black] (descriptionjoin.east)
    -- (description.west |- descriptionjoin.east);

  \coordinate (requestbaseline) at (request.south);
  \coordinate (baselinetarget) at (requestbaseline |- baseline.north);
  \coordinate (guidedtarget) at (description.south |- guided.north);
  \node[inputmerge,draw=ink,text=ink]
    (baselinejoin) at ($(requestbaseline)!.5!(baselinetarget)$) {+};
  \draw[flow,draw=ink]
    (payload.east |- baselinejoin) -- (baselinejoin.west);
  \draw[flow,draw=ink] (requestbaseline) -- (baselinejoin.north);
  \draw[arrow,draw=ink] (baselinejoin.south) -- (baselinetarget);

  \coordinate (guidedentry) at ($(request.south east)+(0,.12)$);
  \node[inputmerge,draw=teal!85!black,text=teal!85!black]
    (guidedjoin) at (description.south |- guidedentry) {+};
  \draw[flow,draw=teal!85!black]
    (request.east |- guidedjoin.west) -- (guidedjoin.west);
  \draw[flow,draw=teal!85!black] (description.south) -- (guidedjoin.north);
  \draw[arrow,draw=teal!85!black] (guidedjoin.south) -- (guidedtarget);
  \node[flowlabel] at ($(baselinejoin)+(.32,0)$) {IR schema design};
  \node[flowlabel,text=teal!85!black] at ($(guidedjoin)+(.32,0)$) {IR schema design};
\end{tikzpicture}%
}
\endgroup

%% file: figures/refiner_two_loops.tex
\begin{tikzpicture}[x=1cm,y=1cm,
  pipelinestep/.style={draw=ink!35,fill=white,line width=0.5pt,
    rounded corners=3pt,font=\sffamily\footnotesize,text width=2.3cm,
    minimum height=1.1cm,align=center,inner sep=2pt},
  twolinestep/.style={minimum height=0.9cm},
  serviceinput/.style={pipelinestep,fill=serviceInputBackground,
    draw=ink!50,rounded corners=9pt},
  generationstage/.style={rectangle split,rectangle split parts=2,
    rectangle split part fill={bluewash,white},
    draw=ink!35,line width=0.5pt,rounded corners=3pt,
    font=\sffamily\footnotesize,text width=2.4cm,align=center,inner sep=2pt,outer sep=0pt},
  loopstep/.style={pipelinestep,text width=2.4cm,outer sep=0pt},
  codesignstep/.style={loopstep},
  flow/.style={-{Stealth[length=1.7mm]},draw=ink!70,line width=0.8pt,
    rounded corners=4pt},
  innerflow/.style={flow,draw=teal},
  outerflow/.style={flow,draw=ochre!90!black},
  note/.style={font=\sffamily\footnotesize,align=center,inner sep=2pt},
  looplabel/.style={font=\sffamily\bfseries\small,align=center,inner sep=0pt},
  loopbadge/.style={looplabel,anchor=west,text=white,rounded corners=2pt,
    inner xsep=4pt,inner ysep=1pt},
  every node/.style={font=\sffamily\footnotesize}]

  \node[looplabel,anchor=north,text=ink] at (6.0,0.08) {RoutineGen Refiner};
  \node[looplabel,anchor=north,text=ink] at (16.5,0.08) {Pet Care Assistant};
  \fill[ochre!8,rounded corners=6pt] (0.0,0.20) rectangle (12.0,5.25);
  \fill[bluewash,rounded corners=6pt] (0.0,0.35) rectangle (2.8,4.50);
  \fill[teal!10,rounded corners=6pt] (2.95,0.35) rectangle (8.95,4.50);
  \node[loopbadge,fill=ochre!70!black] at (0.2,5.0)
    {Human--AI Co-Design Loop};
  \node[loopbadge,fill=teal!85!black] at (3.12,4.21)
    {Automated Refinement Loop};
  \node[loopbadge,fill=ink!80] at (0.2,4.21) {Bootstrapping};

  \node[generationstage,anchor=south] (generate) at (4.35,1.08)
    {\scriptsize\textcolor{ink}{Automated\\Refinement Dataset}
     \nodepart{second}\footnotesize
     \textbf{Generate}\\\textbf{IR payload}\\\textcolor{ink!80}{Service LLM}};
  \coordinate (initialright) at ({1.4cm+1.15cm+2.25pt},0);
  \node[loopstep,minimum height=0.75cm]
    (repair) at (7.5,0 |- generate.two east)
    {\textbf{Analyze}\\\textbf{\& revise}\\\textcolor{ink!80}{Analysis LLM}};
  \coordinate (innercenter) at ($(generate.south)!0.5!(repair.south)$);
  \node[loopstep,twolinestep] (check) at (innercenter |- 0,3.48)
    {\textbf{Translate}\\\textbf{\& Validate}};

  \draw[innerflow] (generate.north) -- (generate.north |- check.west) -- (check.west);
  \draw[innerflow] (check.east) -- (repair.north |- check.east) -- (repair.north);
  \draw[innerflow] (repair.south) -- (7.5,0.73)
    -- (4.35,0.73) -- (generate.south);
  \node[note,font=\sffamily\scriptsize,text=teal!85!black] at (5.95,0.50)
    {Revise prompt, examples, or translator};

  \coordinate (transferlevel) at ($(generate.two west)-(0,0.30)$);
  \node[pipelinestep,twolinestep,outer sep=0pt,anchor=north]
    (initial) at (1.4,3.93)
    {\textbf{Derive routine}\\\textbf{descriptions}};
  \node[pipelinestep,twolinestep,outer sep=0pt,anchor=north]
    (bootschema) at ($(initial.south)-(0,0.235)$)
    {\textbf{Design}\\\textbf{IR schema}};
  \node[pipelinestep,minimum height=1.1cm,outer sep=0pt,anchor=north]
    (bootartifacts) at ($(bootschema.south)-(0,0.235)$)
    {\textbf{Build artifacts}\\[2pt]
     \scriptsize Translator\\[-1.5pt]\scriptsize System prompt\\[-1.5pt]\scriptsize Few-shot examples};
  \node[generationstage,
    rectangle split part fill={bluewash,white},
    anchor=south west] (codesigngenerate)
    at ($(repair.east |- bootartifacts.south)+(generate.two west)-(initialright |- generate.two west)$)
    {\scriptsize\textcolor{ink}{Co-Design Dataset}
     \nodepart{second}\footnotesize
     \textbf{Generate}\\\textbf{IR payload}\\\textcolor{ink!80}{Service LLM}};
  \draw[flow] (initial.south) -- (bootschema.north);
  \draw[flow] (bootschema.south) -- (bootartifacts.north);
  \draw[outerflow] (bootartifacts.east |- transferlevel) -- (generate.two west |- transferlevel);
  \node[codesignstep,twolinestep,
    draw=ochre!85!black,line width=0.8pt]
    (developer) at (codesigngenerate.north |- 0,4.65)
    {\textbf{Developer}\\\textbf{review}};
  \path let \p1=(developer.south), \p2=(codesigngenerate.north),
    \p3=(repair.north), \p4=(repair.south),
    \p5=(check.north), \p6=(check.south),
    \n1={(\y1-\y2-max(1.1cm,\y3-\y4)-(\y5-\y6))/3} in
    node[codesignstep,minimum height=1.1cm,anchor=north]
      (audit) at ($(developer.south)-(0,\n1)$)
      {\textbf{Analyze}\\\textbf{\& propose}\\\textcolor{ink!80}{Analysis LLM}}
    node[codesignstep,twolinestep,anchor=north]
      (codesigncheck) at ($(audit.south)-(0,\n1)$)
      {\textbf{Translate}\\\textbf{\& Validate}};
  \draw[outerflow] (repair.east |- transferlevel) -- (codesigngenerate.two west |- transferlevel);
  \draw[outerflow] (codesigngenerate.north) -- (codesigncheck.south);
  \draw[outerflow] (codesigncheck.north) -- (audit.south);

  \draw[outerflow] (audit.north) -- (developer.south);
  \draw[outerflow] (developer.west) --
    node[pos=0.31,above,note,align=left,inner ysep=0.5pt,font=\sffamily\scriptsize,text=ochre!85!black]
      {Clarify requirements,\\revise data, and resolve conflicts}
    (initial.north |- developer.west)
    -- (1.4,4.50);

  \definecolor{serviceBackground}{HTML}{F8E7EE}
  \definecolor{serviceInputBackground}{HTML}{E5E5E5}
  \filldraw[fill=serviceBackground,draw=ink!25,line width=0.5pt,rounded corners=6pt]
    (15.15,0.20) rectangle (17.85,5.25);
  \node[serviceinput,text width=2.1cm,minimum height=0.65cm,outer sep=0pt]
    (request) at (16.5,4.67) {User request};
  \node[pipelinestep,text width=2.3cm,minimum height=0.65cm,
    outer sep=0pt,anchor=north]
    (planning) at ($(request.south)-(0,0.32)$) {\textbf{Planning}};
  \node[pipelinestep,text width=2.3cm,outer sep=0pt,anchor=north]
    (ir) at ($(planning.south)-(0,0.32)$)
    {\textbf{Generate}\\\textbf{IR payload}\\\textcolor{ink!80}{Service LLM}};
  \node[pipelinestep,text width=2.3cm,minimum height=0.65cm,
    outer sep=0pt,anchor=north]
    (translate) at ($(ir.south)-(0,0.32)$) {\textbf{Translate}};
  \node[note,outer sep=0pt,anchor=north]
    (routine) at ($(translate.south)-(0,0.32)$) {Routine};
  \coordinate (inputmerge) at ($(request.south)!0.5!(planning.north)$);
  \node[serviceinput,text width=2.3cm,minimum height=0.85cm]
    (context) at (13.75,0 |- inputmerge) {Location /\\device\\information};
  \draw[flow,-] (context.east) -- (inputmerge);
  \draw[flow] (request.south) -- (planning.north);
  \fill[ink!70] (inputmerge) circle (1pt);
  \draw[flow] (planning.south) -- (ir.north);
  \draw[flow] (ir.south) -- (translate.north);
  \draw[flow] (translate.south) -- (routine.north);

  \path let \p1=(ir.north), \p2=(translate.south), \n1={\y1-\y2} in
    node[draw=ink!35,fill=black!4,line width=0.5pt,rounded corners=3pt,
      minimum width=2.4cm,minimum height=\n1,inner sep=0pt,
      outer sep=0pt,anchor=north] (library) at (13.75,\y1) {};
  \node[note,font=\sffamily\bfseries\footnotesize,anchor=north]
    at ($(library.north)-(0,0.12)$)
    {Artifacts};
  \node[note,font=\sffamily\scriptsize,text width=2.15cm]
    (generationassets) at ($(library.north)-(0,1.10)$)
    {IR schema\\System prompt\\Few-shot examples};
  \node[note,font=\sffamily\scriptsize,outer sep=0pt]
    (translatorasset) at ($(13.75,0 |- translate.center)-(0,0.05)$) {Translator};
  \coordinate (artifactdivider) at ($(generationassets.south)!0.5!(translatorasset.north)$);
  \draw[ink!25,line width=0.5pt]
    (12.55,0 |- artifactdivider) -- (14.95,0 |- artifactdivider);
  \draw[flow] (developer.east) -- (12.2,0 |- developer.east)
    -- node[pos=0.60,right,note,anchor=south west,font=\sffamily\scriptsize] {Deploy}
    (12.2,0 |- library.west) -- (library.west);
  \draw[flow,dash pattern=on 1.2pt off 1.2pt] (library.east |- ir.center) -- (ir.west);
  \draw[flow,dash pattern=on 1.2pt off 1.2pt] (library.east |- translate.center) -- (translate.west);
\end{tikzpicture}

%% file: appendix.tex
\setlength{\textfloatsep}{10pt plus 2pt minus 2pt}
\makeatletter
\setlength{\@fptop}{0pt}
\setlength{\@fpsep}{20pt}
\setlength{\@dblfptop}{0pt}
\setlength{\@dblfpsep}{20pt}
\makeatother
\section{Routine Descriptions}
\label{sec:description-details}
Each routine description is generated from a user request and its paired routine payload. Payloads supplied for description generation omit the top-level routine name, time-zone fields, and fixed enabled status. Cases are submitted in batches, with a separate description returned for each case. Descriptions are written in English regardless of the request language. User-specified strings, such as notification messages, retain their original language.

Descriptions distinguish conditions that trigger execution from conditions that must hold when execution occurs. They also distinguish current states from state changes and specify whether boundary values are inclusive. They preserve targets, values, units, logical scope, execution order, delays, and refusal semantics. Times and delays are expressed in human-readable terms, while original field names and type structures are excluded. If the request and routine payload differ, the description follows the behavior encoded in the payload, and the meaning of each input and their discrepancy are recorded separately. For schema design, we provide the request and description, but not the discrepancy record. Whether to correct the expected output is decided in the Human--AI Co-Design Loop.

We check the output format and look for missing or duplicate cases. We also check for original field names from a predefined list. Descriptions that fail these checks are regenerated. These are rule-based checks without a separate LLM semantic review, so they do not guarantee that every behavior is faithfully preserved. The later semantic review of the schema is a separate procedure.

\section{Datasets and Location Environments}
\label{sec:dataset-details}
\begin{table*}[!t]
\centering\small
\captionsetup{position=bottom,skip=\papertablesep}
\setlength{\tabcolsep}{5pt}
\renewcommand{\arraystretch}{1.12}
\begin{tabularx}{0.98\textwidth}{@{}Ylrrrrr@{}}
\toprule
Dataset & Location & Devices & \shortstack{Registered\\rooms} & Generation & Refusal & Total \\
\midrule
Automated Refinement & Large & 36 & 4 & 26 & 22 & 48 \\
 & Small & 4 & 0 & 19 & 3 & 22 \\
\midrule
Co-Design & Large & 34 & 5 & 33 & 16 & 49 \\
 & Small & 4 & 0 & 16 & 6 & 22 \\
\midrule
Final Evaluation & Large & 34 & 5 & 60 & 11 & 71 \\
 & Small & 4 & 0 & 18 & 11 & 29 \\
\bottomrule
\end{tabularx}
\caption{Location environments and case counts in each dataset. Devices and registered rooms are counted from the location metadata used in the experiments.}
\label{tab:dataset-locations}
\end{table*}

\subsection{Location Environments}
Each case is associated with a location that provides the available devices and their specifications. We use \emph{large} and \emph{small} to distinguish the size of these device inventories, not physical floor area. Large locations include devices for lighting, climate and air-quality control, sensing, blinds, household appliances, and pet feeding across multiple registered rooms. Small locations contain four devices: a light, an air conditioner, an air purifier, and a television. No rooms are registered in the small-location metadata; this does not mean the location has no physical rooms.

Table~\ref{tab:dataset-locations} lists the environments and case counts. Automated refinement uses one pair of large and small locations. Co-design and final evaluation share another pair of location environments but use different request cases.

\begin{table*}[!t]
\centering\small
\captionsetup{position=bottom,skip=\papertablesep}
\setlength{\tabcolsep}{6pt}
\renewcommand{\arraystretch}{1.18}
\begin{tabularx}{\textwidth}{@{}lYY@{}}
\toprule
 & Generation & Refusal \\
\midrule
User request & Turn on Light1 in living room, between 9 AM and 5 PM when motion is detected & When the external humidity drops below 30\%, turn off the dehumidifier in Kids room \\
\addlinespace
\shortstack[l]{Relevant device\\specification} & MotionSensor1 reports motion; Light1 supports the \texttt{switch:on} command. Both are in Living room. & Dehumidifier2 is in Kids room. Its specification has no \texttt{switch} command for power control. \\
\addlinespace
\shortstack[l]{Expected output\\(summary)} & A routine that turns on Light1 when MotionSensor1 detects motion between 09:00 and 17:00. & A refusal with \texttt{invalid\_request: true}. Reason: \textquotedblleft{}Dehumidifier2 in Kids room does not support switch/power control.\textquotedblright{} \\
\bottomrule
\end{tabularx}
\caption{Generation and refusal examples from the Automated Refinement Dataset. Expected outputs are summarized from the stored payloads.}
\label{tab:dataset-examples}
\end{table*}

\subsection{Generation and Refusal Examples}
\label{sec:dataset-examples}
Table~\ref{tab:dataset-examples} shows two cases from the Automated Refinement Dataset. Both use the large location. Requests are reproduced as stored, and expected outputs are summarized to show the distinction between generation and refusal.

\begin{table}[!ht]
\centering\small
\captionsetup{position=bottom,skip=\papertablesep}
\setlength{\tabcolsep}{3pt}
\renewcommand{\arraystretch}{0.94}
\setlength{\aboverulesep}{1pt}
\setlength{\belowrulesep}{1pt}
\begin{tabularx}{\columnwidth}{@{}Xrr@{}}
\toprule
Feature & \shortstack{Automated\\Refinement} & Co-Design \\
\midrule
Total requests & 70 & 71 \\
Generation / refusal & 45 / 25 & 49 / 22 \\
\midrule
\multicolumn{3}{@{}l}{\textit{Action types}} \\
Conditional execution & 40 & 44 \\
Device commands & 39 & 36 \\
Scheduled execution & 23 & 14 \\
Delay & 5 & 3 \\
Location mode/security change & 2 & 4 \\
Execution limit & 1 & 2 \\
Toggle & 1 & 2 \\
Notification & 8 & 10 \\
Action grouping & 1 & 1 \\
User confirmation & 2 & 2 \\
\midrule
\multicolumn{3}{@{}l}{\textit{Condition operators}} \\
AND & 12 & 10 \\
OR & 2 & 6 \\
Holiday exclusion & 2 & 1 \\
Equality & 27 & 30 \\
Greater than or equal & 9 & 11 \\
Less than or equal & 5 & 4 \\
Inclusive range & 7 & 4 \\
State duration & 1 & 2 \\
\midrule
\multicolumn{3}{@{}l}{\textit{Execution semantics and compositions}} \\
Explicit sequential execution & 5 & 3 \\
Explicit parallel execution & 6 & 5 \\
Explicit execution-time gate & 15 & 11 \\
Presence event or state & 6 & 9 \\
Schedule with conditional execution & 18 & 9 \\
Delay with device commands & 5 & 3 \\
\midrule
\multicolumn{3}{@{}l}{\textit{Expected refusal categories}} \\
Device not found & 2 & 4 \\
Unsupported capability/value & 5 & 5 \\
Unsupported routine feature & 9 & 7 \\
Not a routine request & 5 & 3 \\
Other stored refusal reasons & 4 & 3 \\
\bottomrule
\end{tabularx}
\caption{Feature occurrence in the two development datasets. Generation rows count requests whose expected routine contains the corresponding structure; refusal rows use stored reason codes. Each request is counted at most once per row, and rows overlap. Explicit sequence modes exclude omitted defaults. The table measures occurrence of supported action/condition types and selected semantics, not exhaustive specification or combination coverage, success rates, or population representativeness.}
\label{tab:development-coverage}
\end{table}

\subsection{Development Dataset Coverage}
\label{sec:development-coverage}
We derive the feature categories in Table~\ref{tab:development-coverage} from the supported routine specification and service policy. Only the Automated Refinement and Co-Design Datasets are included. For generation cases, we inspect expected routine payloads and count each case at most once per feature. Location-mode/security actions are distinguished from location-attribute operands. Explicit execution-time gates use \texttt{trigger: Never}; presence cases include events and states. Refusal categories follow the stored \texttt{reason\_code}, including its original assignment of policy refusals. Both datasets contain all supported action and condition types, but this does not establish exhaustive coverage of values, combinations, or natural-language variation.

\section{Bootstrapping and Automated Refinement}
\label{sec:method-details}
\subsection{Validation Criteria}
\label{sec:validation-details}
For generation cases, an exact match of JSON structure and values between the final routine payload and the expected output is the default validation criterion. For fields that need not have a fixed value, such as routine names, the expected output uses an \texttt{\_\_any\_\_} marker to allow any value; a missing field still fails. When multiple outputs are acceptable, a \texttt{\_\_one\_of\_\_} marker lists them, and matching any one is sufficient.

We normalize fields and structures and apply predefined equivalence rules, including equivalence between some sensor sources. Array order is not ignored indiscriminately: sequential actions and command arguments retain their ordering. For refusal cases, we require a refusal and a nonempty reason.

\subsection{IR Schema and Translator Construction}
\label{sec:translator-construction}
For Description-Guided IR, the high-performance LLM receives requests, routine descriptions, and general semantic requirements. These requirements explicitly cover condition roles, targets, values, units, ordering, and refusals, including behaviors that are rare or absent in the cases. The same predefined requirements are also supplied to Baseline IR. The schema specifies both structure and interpretation, including the meaning of omitted fields and mutually exclusive expressions. It also distinguishes triggering conditions from conditions that must hold at execution time, and defines logical scope and ordering. A behavior's expressibility in the IR schema does not imply service support.

A separate LLM call reviews whether the schema can represent the cases and general requirements. It must identify the fields and interpretation rules supporting its judgments. We also check whether distinct behaviors can be expressed as different IR payloads. Unrepresentable semantics and ambiguous interpretations are fed back to revise the schema. The original data structure is withheld during this review as well.

Translator generation additionally receives routine payloads and specifications. Code is generated in parts and checked for executability. Routine payloads are then reverse-converted into IR payloads and translated back into routines to check for information loss. Failed round-trip conversions are recorded for refinement, but passing this check on every case is not required to finalize the schema. We next review requests, reverse-converted IR payloads, and the IR schema together, without supplying routine payloads or translation code. If the structure is sufficient, field descriptions are revised first; if not, the schema structure is revised.

\begin{table*}[!t]
\centering\small
\captionsetup{position=bottom,skip=\papertablesep}
\begin{tabularx}{0.94\textwidth}{@{}Yl@{}}
\toprule
Task & Reasoning level \\
\midrule
Derive routine descriptions & high \\
Design candidate IR schemas & medium \\
Review schema coverage; review reverse-converted IR payloads & high \\
Plan, generate in parts, and repair the initial translator & low \\
Revise schema field descriptions & low \\
Review revised field descriptions & high \\
Write the initial system prompt and select few-shot examples from the Automated Refinement Dataset & high \\
Reselect few-shot examples during bootstrapping (system prompt unchanged) & low \\
Analyze errors and revise artifacts in the Automated Refinement Loop & high \\
Analyze Co-Design Dataset failures and propose revisions & high \\
\bottomrule
\end{tabularx}
\caption{Task-specific reasoning levels of the analysis LLM (Claude Opus 5.5) for bootstrapping, the Automated Refinement Loop, and Co-Design Dataset analysis in Description-Guided IR.}
\label{tab:design-model-settings}
\end{table*}

After confirming that all issues from the required semantic review have been resolved, we construct a system prompt and few-shot examples. IR payloads in few-shot examples for routine generation are translated and compared with expected outputs; refusal examples are checked for the required refusal format. We check that the library meets its mandatory requirements and save the initial version. The schema is then fixed and the Automated Refinement Loop begins.

For both IR methods, the service LLM receives a compact device context containing device identities, capabilities, attributes, and available choices. Execution details such as numeric ranges and value-type annotations are omitted from this context. The translator receives the generated IR payload and detailed device specifications. It resolves commands, units, and types, checks supported operations, permitted values, and ranges, and constructs a routine that conforms to the routine specification.

\subsection{Diagnosis and Revision in the Automated Refinement Loop}
The high-performance LLM receives representative cases for each failure type, pass records for recent candidates, and the rationale and outcome of the previous revision. It first selects one revision target and provides a rationale, then revises it in a separate call. Missing semantics that the schema can express prompt a review of the system prompt or few-shot examples; incorrect translation of a semantically correct IR payload prompts a review of the translator. We check translator syntax and required functions, as well as required input placeholders in the system prompt, before saving a new candidate. The same service LLM then processes the entire Automated Refinement Dataset again, including checks for regressions on previously passing cases. The candidate with the most passing cases is retained. Even if an expected output or schema is diagnosed as problematic, the loop does not automatically change it or exclude the case. Such issues are reviewed later in the Human--AI Co-Design Loop.

\subsection{Model Settings for Bootstrapping and Refinement}
\label{sec:design-model-settings}
Claude Opus 5.5 performed bootstrapping and the analysis and revision steps of the Automated Refinement Loop. Reasoning levels refer to Claude's effort setting \citep{anthropic-effort} and vary by task even though the model is the same. Table~\ref{tab:design-model-settings} lists settings used for bootstrapping and the Automated Refinement Loop in Description-Guided IR. Baseline IR and Direct Generation use the same reasoning levels for the tasks they share; Direct Generation has no schema or translator tasks. Model outputs for user requests, in both refinement and final evaluation, are generated by the service model and reasoning level assigned to each experimental condition.

Few-shot example reselection during bootstrapping is a recovery step that rebuilds only the few-shot examples while leaving the system prompt unchanged. When there are at most five candidate few-shot examples, all are selected without an additional model call.

\section{Offline Evaluation Details}
\label{sec:details}
\begin{table*}[!t]
\centering\small
\captionsetup{position=bottom,skip=\papertablesep}
\setlength{\tabcolsep}{4pt}
\begin{tabularx}{\textwidth}{@{}lXrrrrr@{}}
\toprule
Model & Method & P & E & Pass & Stale & Stop \\
\midrule
Flash-Lite/medium & Direct Generation & 14 [8--19] & 14 [8--19] & 0 & 0 & 3 \\
Flash-Lite/medium & Baseline IR & 15 [14--28] & 16 [15--29] & 1 & 2 & 0 \\
Flash-Lite/medium & Description-Guided IR & 13 [10--17] & 11 [9--18] & 2 & 0 & 1 \\
Flash/low & Direct Generation & 14 [11--18] & 15 [12--19] & 2 & 1 & 0 \\
Flash/low & Baseline IR & 12 [7--22] & 11 [8--23] & 1 & 1 & 1 \\
Flash/low & Description-Guided IR & 12 [3--23] & 13 [4--24] & 2 & 1 & 0 \\
Flash/minimal & Direct Generation & 14 [11--14] & 14 [11--15] & 1 & 0 & 2 \\
Flash/minimal & Baseline IR & 12 [6--29] & 13 [7--24] & 2 & 1 & 0 \\
Flash/minimal & Description-Guided IR & 14 [9--18] & 15 [10--19] & 2 & 1 & 0 \\
\bottomrule
\end{tabularx}
\caption{Recorded proposals and evaluations: median [minimum--maximum] across three independently developed libraries per row. Pass, Stale, and Stop count runs by termination reason. All counts concern the retained workflow attempts.}
\label{tab:recorded-budget-summary}
\end{table*}

\begin{table*}[!t]
\centering\small
\captionsetup{position=bottom,skip=\papertablesep}
\begin{tabular}{@{}llrrr@{}}
\toprule
Model & Method & Run 1 & Run 2 & Run 3 \\
\midrule
Flash-Lite/medium & Direct Generation & 82.00 & 86.00 & 86.00 \\
 & Baseline IR & 92.00 & 99.00 & 94.00 \\
 & Description-Guided IR & 98.00 & 99.00 & 97.00 \\
\addlinespace
Flash/low & Direct Generation & 85.00 & 83.00 & 91.00 \\
 & Baseline IR & 96.00 & 96.00 & 95.00 \\
 & Description-Guided IR & 98.00 & 95.00 & 95.00 \\
\addlinespace
Flash/minimal & Direct Generation & 84.00 & 87.00 & 91.00 \\
 & Baseline IR & 92.00 & 91.00 & 91.00 \\
 & Description-Guided IR & 92.00 & 91.00 & 93.00 \\
\bottomrule
\end{tabular}
\caption{Accuracy (\%) on the Final Evaluation Dataset for all 27 runs in the main comparison.}
\label{tab:accuracy-runs}
\end{table*}

\begin{table}[!ht]
\centering\small
\captionsetup{position=bottom,skip=\papertablesep}
\begin{tabular}{@{}lr@{}}
\toprule
Model & \shortstack{Regular-price cost\\(USD / 1,000 requests)} \\
\midrule
Flash-Lite/medium & 7.17 \\
Flash/low & 23.45 \\
Flash/minimal & 21.35 \\
\bottomrule
\end{tabular}
\caption{Estimated Description-Guided IR cost from token usage in successful calls for the 100 cases in the Final Evaluation Dataset. Calculations use unrounded mean token counts.}
\label{tab:cost}
\end{table}

\subsection{Generation Settings and Stopping Criteria}
\label{sec:eval-details}
All methods use temperature 0.1, a fixed seed of 0, a maximum output length of 32,767 tokens, and 10 concurrent requests. A request permits up to five model-call attempts, with retries only after a call failure. Generated outputs are not retried based on correctness, and expected outputs are not used to decide retries or select among generated responses. We perform at most 60 evaluations of refinement candidates in addition to the initial evaluation. Refinement stops after 10 consecutive evaluations without improvement once the best accuracy on the Automated Refinement Dataset reaches at least 90\%. It also stops when all cases pass or a revision termination condition is met. Direct Generation uses five few-shot examples; the IR methods use up to five validated few-shot examples.

\begin{table*}[!t]
\centering\small
\captionsetup{position=bottom,skip=\papertablesep}
\setlength{\tabcolsep}{5pt}
\begin{tabular}{@{}llrrrrr@{}}
\toprule
Model & Method & \shortstack{Input\\tokens} & \shortstack{Output\\tokens} & \shortstack{Reasoning\\tokens} & \shortstack{Generation\\(78)} & \shortstack{Refusal\\(22)} \\
\midrule
Flash-Lite/medium & Direct Generation & 16,188.48 & 120.60 & 1,723.60 & 82.05 & 93.94 \\
 & Baseline IR & 13,563.08 & 101.13 & 1,158.16 & 94.44 & 96.97 \\
 & Description-Guided IR & 13,485.41 & 103.78 & 1,146.19 & 97.86 & 98.48 \\
\addlinespace
Flash/low & Direct Generation & 15,466.81 & 124.09 & 762.73 & 83.33 & 96.97 \\
 & Baseline IR & 13,647.75 & 108.37 & 426.89 & 94.87 & 98.48 \\
 & Description-Guided IR & 12,326.41 & 101.53 & 559.94 & 94.87 & 100.00 \\
\addlinespace
Flash/minimal & Direct Generation & 16,987.81 & 121.89 & 0.00 & 85.90 & 92.42 \\
 & Baseline IR & 13,579.41 & 105.06 & 0.00 & 91.45 & 90.91 \\
 & Description-Guided IR & 13,707.08 & 105.74 & 0.00 & 93.16 & 87.88 \\
\bottomrule
\end{tabular}
\caption{Mean tokens per request and accuracy (\%) for generation and refusal cases in the Final Evaluation Dataset. Values are averaged over three runs per service model and method.}
\label{tab:tokens}
\label{tab:splits}
\end{table*}

\begin{table}[!ht]
\centering\small
\captionsetup{position=bottom,skip=\papertablesep}
\setlength{\tabcolsep}{4pt}
\begin{tabularx}{\columnwidth}{@{}Yrrr@{}}
\toprule
Run & Initial (\%) & Refined (\%) & Gain (pp) \\
\midrule
1 & 44.00 & 98.00 & +54.00 \\
2 & 95.00 & 99.00 & +4.00 \\
3 & 88.00 & 97.00 & +9.00 \\
\bottomrule
\end{tabularx}
\caption{Accuracy before and after the Automated Refinement Loop for Description-Guided IR with Flash-Lite/medium, evaluated on the 100 cases in the Final Evaluation Dataset.}
\label{tab:initial-improvement}
\end{table}

\subsection{Recorded Refinement Budgets}
\label{sec:recorded-budgets}
Table~\ref{tab:recorded-budget-summary} summarizes archived records for the 27 main runs. P counts analysis/revision proposals, including decisions that produced no candidate. E counts complete evaluations of the 70-case Automated Refinement Dataset, including the initial evaluation. Subtracting the initial evaluation from E gives the number of evaluated revisions; multiplying E by 70 gives case evaluations, not provider calls. Bootstrapping checks and final evaluation are excluded. Equal maximum budgets do not imply equal realized search effort.

Pass denotes all 70 development cases passing. Stale denotes ten consecutive candidate evaluations without improving the best score after reaching at least 90\%. Stop denotes an explicit analysis-model stop decision, not proof that further improvement was impossible. Across the main runs, 13 ended with Pass, seven with Stale, and seven with Stop; none reached the 60-revised-evaluation limit. The evaluation-count medians for Baseline and Description-Guided IR were respectively 16 and 11 with Flash-Lite/medium, 11 and 13 with Flash/low, and 13 and 15 with Flash/minimal. The records therefore do not establish a consistent reduction in evaluations with Description-Guided IR.

The three additional Opus-only development runs used 6--12 evaluations including the initial one. Two reached Pass. Run \texttt{8946760/4} stopped after ten consecutive proposals without a usable candidate, recorded as \texttt{data\_conflict}. It made 21 proposals but evaluated only 11 revisions (12 evaluations in total). The best saved candidate, proposal 9 at 69/70 development cases, was retained; the workflow completed with warnings. Re-evaluating these same three libraries with Flash-Lite adds no development runs.

The refinement statistics in this appendix were derived from saved execution and candidate-evaluation records. Selected candidates are identified by workflow run, attempt, and candidate hash; repeated metadata versions and equal run ordinals across methods do not identify paired library construction. The records include token usage but no actual development billing totals.

\subsection{Accuracy by Run}
Table~\ref{tab:accuracy-runs} reports accuracy on the 100 cases in the Final Evaluation Dataset for each run.

\begin{table*}[!t]
\centering\small
\captionsetup{position=bottom,skip=\papertablesep}
\begin{tabularx}{\textwidth}{@{}lYY@{}}
\toprule
 & Baseline IR & Description-Guided IR \\
\midrule
IR schema & \texttt{Operation.value} refers to \texttt{Value}, whose string form is \texttt{\{"text": <string>\}}. & \texttt{Step.value} accepts a scalar, array, or object. \\
IR payload & \texttt{\{"op":"set", "capability":"airConditionerMode", "value":"dry"\}} & \texttt{\{"kind":"device", "op":"set", "capability":"airConditionerMode", "attribute":"airConditionerMode", "value":"dry"\}} \\
Translation & Raises \texttt{ValueError: Value must be an object: 'dry'}. No routine is produced. & Emits \texttt{setAirConditionerMode} with argument \texttt{\{"string":"dry", "type":"string"\}} under the daily 17:00 schedule. \\
Evaluation & Fails due to translation error. & Passes the stored validator, including its normalizations. \\
\bottomrule
\end{tabularx}
\caption{Recorded IR schema requirements, generated IR payload excerpts, and outcomes for the same request in two independently developed libraries.}
\end{table*}

\begin{table*}[!t]
\centering\small
\captionsetup{position=bottom,skip=\papertablesep}
\setlength{\tabcolsep}{7pt}
\begin{tabular}{@{}llrrrrr@{}}
\toprule
Development & \shortstack[l]{Generation\\LLM} & Run & Total / 100 & Generation / 78 & Refusal / 22 & \shortstack{Latency\\(s)} \\
\midrule
Opus only & Opus & 1 & 98 & 77 & 21 & 3.59 \\
 & & 2 & 99 & 77 & 22 & 3.74 \\
 & & 3 & 95 & 74 & 21 & 4.15 \\
\addlinespace
Opus only & Flash-Lite & 1 & 91 & 71 & 20 & 3.28 \\
 & & 2 & 94 & 72 & 22 & 4.08 \\
 & & 3 & 95 & 75 & 20 & 6.73 \\
\addlinespace
Opus + Flash-Lite & Flash-Lite & 1 & 98 & 76 & 22 & 4.43 \\
 & & 2 & 99 & 77 & 22 & 4.56 \\
 & & 3 & 97 & 76 & 21 & 5.04 \\
\bottomrule
\end{tabular}
\caption{Passed cases and mean request latency for Description-Guided IR under each combination of development and generation LLMs. Generation LLM identifies the model used to generate IR payloads during final evaluation. Opus denotes Claude Opus 5.5; Flash-Lite denotes Gemini 3.5 Flash-Lite. Matching run numbers in the first two combinations use the same library. The last combination reuses results from the main comparison.}
\label{tab:cross-model-runs}
\end{table*}

\subsection{Request Processing Cost at Regular Prices}
\label{sec:cost-details}
We used Google Cloud Global Standard regular prices checked on October 12, 2026 \citep{google-pricing}. Input/output prices per million tokens are \$1.50/\$7.50 for Flash and \$0.30/\$2.50 for Flash-Lite. Output billing includes reasoning tokens. Flash's regular prices are the published prices effective January 1, 2027.
We applied the input price to mean input tokens and the output price to the sum of mean response-output tokens and mean reasoning tokens. Each count was divided by one million and multiplied by its corresponding price. We summed the amounts and multiplied by 1,000 to estimate the cost of 1,000 requests (Table~\ref{tab:cost}).
Flash-Lite/medium was 69.4\% and 66.4\% less expensive than Flash/low and Flash/minimal, respectively.

\begin{samepage}
\textit{Promotion.} Flash has a 50\% credit promotion through December 31, 2026, giving effective input/output prices of \$0.75/\$3.75. The corresponding cost per 1,000 requests is \$11.73 for low and \$10.68 for minimal. Costs and savings in the main text use regular prices without the promotion. Estimates exclude cache discounts, failed calls without usage records, and library development costs, and therefore differ from actual bills.
\end{samepage}

\subsection{Token Usage and Accuracy by Expected Outcome}
For each run, we summed token counts recorded for successful calls over the 100 cases in the Final Evaluation Dataset and divided by the number of requests. We then averaged the values over three independent runs. Output token counts refer to responses before translation and exclude reasoning tokens. Reasoning tokens were computed by subtracting input and response-output tokens from total tokens (Table~\ref{tab:tokens}).

\subsection{Effect of Automated Refinement}
\label{sec:initial-improvement-details}
Table~\ref{tab:initial-improvement} compares the initial and refined libraries from three runs. Each initial library is the result of bootstrapping, before the Automated Refinement Loop begins. Initial accuracy ranged widely, from 44\% to 95\%, but all refined libraries reached 97--99\%. The Automated Refinement Loop thus produced consistently high-quality libraries despite variation in initial quality.

\subsection{Recorded Value-Representation Case}
\label{sec:recorded-value-case}
Final-evaluation case 52 asks to set the air conditioner to dry mode daily at 17:00. The excerpts are from Baseline IR run \texttt{8944892/2} and Description-Guided IR run \texttt{8944850/1}, both using Flash-Lite/medium. Unrelated fields are omitted.

The Baseline output carries the requested mode but violates its schema's value wrapper; the translator rejects it. The Description-Guided schema permits that scalar representation and its translator constructs the expected routine. Across all three libraries per method, this request passes in two Baseline and three Description-Guided libraries. This is a diagnostic example, not an additional experiment or an isolated estimate of the effect of schema-design inputs.

\section{Role of the Service LLM in Refinement (Detailed Results)}
\label{sec:cross-model-details}
Table~\ref{tab:cross-model-runs} provides per-run results for the Description-Guided IR comparison in Table~\ref{tab:cross-model}. Claude Opus 5.5 performed bootstrapping, analysis, and revision in all conditions, with task-specific reasoning levels listed in Appendix~\ref{sec:design-model-settings}. In the Opus-only development condition, Opus 5.5/low also generated IR payloads for user requests. In the condition using both Opus and Flash-Lite, Flash-Lite/medium generated those payloads. During final evaluation, the model indicated in each row generated IR payloads. The translator converted them into final routines for evaluation by the common validator.

For each run, request latency was averaged over the 100 requests, including refusals and failures, using the measurement scope defined in Section~\ref{sec:experimental-settings}. It measures routine generation by the LLM indicated in each row, not validation against expected outputs.

Evaluating the libraries developed with Opus using Flash-Lite produced 20 failures across the three runs, including final routine payload mismatches, translation exceptions, and missed refusals.